\documentclass[11pt,a4paper]{article}
\usepackage{emnlp2021}
\usepackage{times}
\usepackage{latexsym}

\usepackage{url}
\usepackage{algorithm}
\usepackage[normalem]{ulem}
\usepackage[noend]{algpseudocode}
\usepackage{bbm}
\usepackage{dsfont}
\usepackage{setspace}

\usepackage{amsmath,amsfonts,bm}
\usepackage{fixltx2e}
\usepackage{multirow}
\usepackage{tabularx}
\usepackage{caption}
\usepackage{wrapfig}
\usepackage[export]{adjustbox}
\usepackage{xspace}
\usepackage{kotex}
\usepackage{graphicx}
\usepackage{microtype}
\usepackage{graphicx}
\usepackage{subfigure}
\usepackage{soul}
\usepackage{booktabs} 

\usepackage{hyperref}
\usepackage{enumitem}
\usepackage{algorithm}
\usepackage{xcolor}
\usepackage[normalem]{ulem}
\usepackage[noend]{algpseudocode}

\usepackage{caption}
\definecolor{auburn}{rgb}{0.43, 0.21, 0.1}
\definecolor{ao(english)}{rgb}{0.0, 0.5, 0.0}

\newcommand{\ie}{{\em i.e.,}\xspace}

\def\eqref#1{equation~\ref{#1}}

\def\1{\bm{1}}

\DeclareMathAlphabet{\mathsfit}{\encodingdefault}{\sfdefault}{m}{sl}
\SetMathAlphabet{\mathsfit}{bold}{\encodingdefault}{\sfdefault}{bx}{n}

\def\gD{{\mathcal{D}}}

\def\gX{{\mathcal{X}}}
\def\gY{{\mathcal{Y}}}

\DeclareMathOperator*{\argmin}{arg\,min}

\usepackage[nameinlink]{cleveref}
\crefformat{section}{\S#2#1#3} 
\crefformat{subsection}{\S#2#1#3}
\crefformat{subsubsection}{\S#2#1#3}

\usepackage{soul,xcolor}

\title{Nearest Neighbour Few-Shot Learning for Cross-lingual Classification}
\author{M Saiful Bari \thanks{\ \ Work done while Saiful was interning at Amazon AI} $ \ ^{\S \P}$, Batool Haider $^{\S}$, Saab Mansour$^{\S}$  \\
  $^\S$ Amazon AI \\
  $^\P$ Nanyang Technological University, Singapore \\
  $^\P$\texttt{bari0001@e.ntu.edu.sg}\\ $^\S$\texttt{\{bhaider, saabm\}@amazon.com } \\}

\date{}

\usepackage{todonotes}
\newcounter{todocounter}

\begin{document}

\setstcolor{blue}

\maketitle
\begin{abstract}

    Even though large pre-trained multilingual models (e.g. mBERT, XLM-R) have led to significant performance gains on a wide range of cross-lingual NLP tasks, success on many downstream tasks still relies on the availability of sufficient annotated  data. Traditional fine-tuning of pre-trained models using only a few target samples can cause over-fitting. This can be quite limiting as most languages in the world are under-resourced. In this work, we investigate cross-lingual adaptation using a simple nearest neighbor few-shot ($<15$ samples) inference technique for classification  tasks. We experiment using a total of 16 distinct languages across two NLP tasks- XNLI and PAWS-X. Our approach consistently improves traditional fine-tuning using only a handful of labeled samples in target locales. We also demonstrate its generalization capability across tasks.  
    
   
\end{abstract}

\section{Introduction}  \label{sec:intro}



The rise of massively pre-trained multilingual language models (LM)\footnote{We loosely use the term LM to describe unsupervised pretrained models including Masked-LMs and Causal-LMs
} \cite{XLM,XLMR,chi2020infoxlm,luo2020veco,xue2020mt5} has significantly improved cross-lingual generalization across many languages \cite{xBERT,pires-etal-2019-multilingual,k2020crosslingual,Keung_2019}. Recent work on zero-shot cross-lingual adaptation \cite{fang2020filter,pfeiffer2020madx,bari2020multimix}, in the absence of labelled target data, has also demonstrated impressive performance gains. Despite these successes, however, there still remains a sizeable gap between supervised and zero-shot performances. On the other hand, when limited target language data are available (i.e few-shot setting), traditional fine-tuning of large pre-trained models can cause over-fitting \cite{overfitting}.

\noindent One way to deal with the scarcity of annotated data is to augment synthetic data using techniques like paraphrasing \cite{gao2020paraphrase}, machine translation \cite{BT} and/or data-diversification \cite{bari2020multimix}. Few-shot learning, on the other hand, deals with handling out-of-distribution (OOD) generalization problems using only a small amount of data \cite{Koch2015SiameseNN,vinyals2017matching,snell2017prototypical,santoro2017simple,finn2017modelagnostic}. In this setup, the model is evaluated over few-shot tasks, such that the model learns to generalize to new data (query set) using only a hand full of labeled samples (support set). 

\noindent In a cross-lingual few-shot setup, the model learns cross-lingual features to generalize to new languages. Recently, \citet{nooralahzadeh2020zeroshot} used \emph{Meta-Learning} \cite{MAML} for few-shot adaptation on several cross-lingual tasks. Their few-shot setup used full development datasets of various target languages (XNLI development set, for instance, has over 2K samples). In general, they showed the effectiveness of cross-lingual meta-training in the presence of a large quantity of OOD data. However, they did not provide any fine-tuning baseline. On the contrary, \cite{lauscher2020zero} explored few-shot learning but did not explore beyond fine-tuning. To the best of our knowledge, there has been no prior work in cross-lingual NLP that uses only a handful of target samples ($<15$) and yet surpasses or matches traditional fine-tuning (on the same number of samples).



 \noindent Traditional finetuning (parametric) approaches require proper hyperparameter tuning techniques for the learning rate, scheduling, optimizer, batch size, up-sampling few-shot support samples and failing to do so would often led to model over-fitting. It can  be expensive  to  update parameters of large  model frequently for few shot adaption, each time there is a fresh batch of support samples. As the model grows bigger, it becomes almost unscalable to update weights frequently for few shot adaptation. It takes significant amount of time to update gradients for a few number of samples and then perform inference. \\

\noindent In this work, we explore a  simple \emph{Nearest Neighbor Few-shot Inference} (NNFS) approach for cross-lingual classification tasks. Our main objective is to utilize very few samples to perform adaptation on a given target language. To achieve this, we first fine-tune a multilingual LM on a high resource source language (\ie English), and then apply few-shot inference using few support examples from the target language. Unlike other popular meta-learning approaches that focus on improving the fine-tuning/training setup to achieve better generalization \cite{ finn2017modelagnostic, Ravi2017}, our approach applies to the inference phase. Hence, we do not update the weights of the  LM using target language samples. This makes our approach complimentary to other regularized fine-tuning based few-shot meta-learning approaches. Our key contributions are as follows:
\begin{itemize}[leftmargin=*]
\itemsep0em 
    \item We propose a simple method for cross-lingual few-shot adaptation on classification tasks during inference. Since our approach applies to inference, it does not require updating the LM weights using target language data.
    
    \item Using only a few labeled target support samples, we test our approach across 16 distinct languages belonging to two NLP tasks (XNLI and PAWS-X), and achieve consistent sizable improvements over traditional fine-tuning. 
    \item We also demonstrate that our proposed method can generalize well not only across languages but also across tasks. 
    \item As the support sets are minimal in size, subsequent results obtained using them can suffer from high variability. We borrow the idea of \textit{episodic testing} widely used in computer vision few-shot tasks, to evaluate few-shot performance for NLP tasks (more details in section \ref{sec:eval_setup}).
\end{itemize}


\section{Method}  \label{sec:method}
 The objective of few-shot learning is to adapt from a source distribution to a new target distribution using only few samples. The traditional few-shot setup \cite{finn2017modelagnostic,proto_net,vinyals2017matching} involves adapting a model to the distribution of new classes. Similarly, in a cross-lingual setup, we adapt a pre-trained LM, that has been fine-tuned using a high resource language, to a new target language distribution~\cite{lauscher2020zero,nooralahzadeh2020zeroshot}.

 \subsection{Setup} 
 We begin by fine-tuning a pre-trained model $\theta_{lm}$ \cite{XLMR} to a specific task $\mathcal{T}_s$ using a high resource (source) language data set $\mathcal{D}_{src}=(\mathcal{X}_{src}, \mathcal{Y}_{src})$, to get an adapted model $\theta_{\mathcal{T}_s}^{src}$. We use $\theta_{\mathcal{T}_s}^{src}$ to perform few-shot adaptation. 
 
 
\noindent In our few-shot setup, we assume to possess very few labeled support samples  $\mathcal{D}_{s}=(\mathcal{X}_{s}, \mathcal{Y}_{s})$ from the target language distribution. A support set covers $C$ classes, where each class carries $N$ number of samples. This is a standard \textbf{\emph{$C$-way-$N$-shot}} few-shot learning setup. The objective of our proposed method is to classify the unlabeled query samples $\mathcal{D}_q=(\mathcal{X}_q$). We denote the latent representation of the support and query samples as $X_s$ and $X_q$, respectively, where $X_s=\theta_{\mathcal{T}_s}^{src}(\mathcal{X}_s)$ and $X_q=\theta_{\mathcal{T}_s}^{src}(\mathcal{X}_q)$ .
 
 \subsection{Nearest Neighbor Class} 
 
 \noindent Let $|\mathcal{D}_s|$ and $|\mathcal{D}_q|$ be the total number of support and query samples. For query samples $\mathcal{X}_q$, feature representations $X_q$ is obtained by forward propagation on $\theta_{\mathcal{T}_s}^{src}$ model. For each query representation $x_q$, we define a latent binary assignment vector $y_q = [y_{q,1}, y_{q,2}. . . , y_{q,C}]$. Here, $y_{q,i}$ is a binary variable such that,
 
 \vspace{-1em}

 \begin{equation}
    y_{qi} = \mathds{1}_{i}(y_{q}) :=
\begin{cases}
1 &\text{if } y_q \in i, \\
0 &\text{if } y_q \notin i.
\end{cases} 
\end{equation}
and $\sum_i y_{qi} = 1$. Let $\mathcal{Y}_q$ denote the $\mathcal{R}^{N_q \times C}$ matrix where each row represents the $y_q$ term of each query. 

\noindent We compute the centroid, $m_c$, of each class by taking the mean of its support representations $\mathcal{X}_s$. Next, we compute the distances between each $x_q$ and $m_c$ (Equation \ref{eq:knn}). Our loss function becomes,


\begin{equation}
    \sum_{i=1}^{N_q}\sum_{c=1}^{C} y_{q,c} d(x_q, m_c)
    \label{eq:knn} 
\end{equation}

\noindent Finally, we assign each $x_q$ the label of the class it has the minimum distance to. This is done using the following function,

\vspace{-1em}
\begin{equation}
        y_{qc^{*}} = 
            \begin{cases}
            1 &\text{if } c =\underset{\rm c^{*}\in\{1,2,...,C\}}{\rm \argmin}  d(x_q, m_c) \\
            0 & otherwise
            \end{cases} 
    \label{eq:knn_optim} 
\end{equation}

\begin{algorithm}[H]
\footnotesize
\caption{\small{Nearest Neighbor Few-shot Inference}}
\label{alg:FewShotAdaptation}
 \textbf{Input:} Model $\theta_{\mathcal{T}_s}^{src}$ trained using source language, support set $\gD_s=(\gX_s,\gY_s)$, query Set $\gD_q=(\gX_q$), mean representation of train/dev samples $m_s$\\
 \textbf{Output:} Distribution of the query label, $\gY_q$ 
\begin{algorithmic}[1]
    
    \setstretch{1}
    \State \texttt{/* feature representation normalization */}
    
    \setstretch{1}
    \State $\hat{X_s}, \hat{X_q} = \theta_{\mathcal{T}_s}^{src}(\gX_s), \theta_{\mathcal{T}_s}^{src}(\gX_q)$
    \State $\tilde{X_s}, \tilde{X_q} = \hat{X_s}-m_s, \hat{X_q}-m_s$
    \State $X_s, X_q = \frac{\tilde{X_s}}{||\tilde{X_s}||_2}, \frac{\tilde{X_q}}{||\tilde{X_q}||_2}$
    \State \texttt{/* Calculate $m_c$ */}
    \State $\eta = \frac{1}{|\mathcal{D}_s|}\sum_{i=1}^{|\mathcal{D}_s|} X_s - \frac{1}{|\mathcal{D}_q|}\sum_{i=1}^{|\mathcal{D}_q|} X_q$
    \State $X_q = X_q + \eta$
    
    \State\setstretch{1} \texttt{/* Calculate mean representation of each of the classes */}
    
    \setstretch{1}
    \State $\hat{m}_c =  \frac{1}{\sum_{i}^{|\mathcal{D}_s|} \mathds{1}_{\{c\}}(\mathcal{Y}_i)} \sum_{i}^{|\mathcal{D}_s|} \mathds{1}_{\{c\}}(\mathcal{Y}_i) X_i$
    \State $\hat{y}_{qc^{*}} = \begin{cases}
            1 &\text{if } c^{*} =\underset{\rm c\in\{1,2,...,C\}}{\rm \argmin}  (1-cos(X_q, \hat{m}_c)) \\
            0 & otherwise
            \end{cases}$

    \State\setstretch{1} \texttt{/*$\mathcal{D}_{i}^c$ accumulates all the samples for the class $c$ from $i$ dataset */}
    
    \setstretch{1}
    \State $m_c = \frac{1}{|\mathcal{D}_s^c|+|\mathcal{D}_q^c|} \sum_{\textsc{x}^{\prime} \in \{X_s^c, X_q^c\}} \frac{\exp({\cos({{\textsc{x}^\prime}, \hat{m_c}})})}{\sum_{c=1}^c\exp({\cos({\textsc{x}^\prime}, \hat{m_c}})}\textsc{x}^{\prime}$ 
    \State $a_q=[a_{q,1}, a_{q,2}, ... , a_{q,C}] ; a_{q,c} = d(x_q, m_c)$
    \State $y_q^i = \frac{exp(-a_q)}{exp(-a_q)[1,1,...]^t}$
    \State Return $\mathcal{Y}_q = \Big\{y_q\Big\}_{q=1}^{|\mathcal{X}_q|}$
    
\end{algorithmic}
\end{algorithm}

Traditional \emph{inductive inference} handles each query sample (one at a time), independent of other query samples. On the contrary, our proposed approach includes additional \emph{Normalization} and \emph{Transduction} steps. Algorithm \ref{alg:FewShotAdaptation} illustrates our approach. Here we discuss these additional steps in  more detail. \\

\noindent\textbf{Norm.} We measure the cross-lingual shift as the difference between the mean representations of the support set (target language) and the training set (en), $m_s$. We then perform cross-lingual shift correction on the query set. To achieve this, at first, we extract the latent representation of both support and query samples from $\theta_{\mathcal{T}_s}^{src}(\gX_s)$. We then  center the representation (Alg \ref{alg:FewShotAdaptation} \#3) by subtracting the mean representation of the train/dev data of the source language, followed by L2 normalization of both representations (train/dev). Algorithm \ref{alg:FewShotAdaptation} (\#2-7) further details our approach. \\ 

\noindent\textbf{Transduction.} We apply prototypical rectification (\textit{proto-rect}) \cite{proto-rect} on the extracted features of LM. In the rectification step, to compute $m_c$ (in Alg.\ref{alg:FewShotAdaptation}), initially, we obtain the mean representation for each of the support classes by taking the weighted combination of $X_s$ and $X_q$. Finally, we calculate predictions on the query set using equation \ref{eq:knn_optim}. We also present our proposed NNFS inference in Figure 2 in the Appendix.

\section{Experimental Settings} \label{sec:exps}

\subsection{Data}
\noindent We use two standard multilingal datasets - \emph{\textbf{XNLI }} \cite{multiNLI} (15 languages) and \emph{\textbf{PAWS-X}} \cite{zhang-etal-2019-paws} (7 languages) to evaluate our proposed method. Additional details on languages and complexity of the task can be found in the Appendix. For few-Shot inference, we use samples from the target language development data to construct the support sets and the test data to construct the query sets. 



\begin{table*}[h!]
    \resizebox{1\linewidth}{!}{%
        \begin{tabular}{lcccccccccccccccc}
            \toprule 
            Exp. Type & Resource & \textbf{fr} & \textbf{es} & \textbf{de} & \textbf{el} & \textbf{bg} & \textbf{ru} & \textbf{tr} & \textbf{ar} & \textbf{vi} & \textbf{th} & \textbf{zh} & \textbf{hi} & \textbf{sw} & \textbf{ur} & avg\\
            \midrule
            \midrule
            \multicolumn{17}{c}{$\theta_{\mathcal{T}_s}^{src}$  = Finetuned-XLM-R-large with XNLI dataset} \\
            \midrule
            \midrule
Zero-Shot &  en &  83.1 &84.8 &83.0 &82.2 &83.4 &80.1 &78.8 &78.8 &80.1 &78.1 &79.4 &76.7 &72.7 &72.9  & 79.6\\
NN &  en+fs-3.5 &  83.0 &84.6 &82.7 &82.0 &83.3 &80.3 &78.9 &79.2 &80.2 &78.3 &79.5 &76.6 &71.9 &73.0  & 79.5\\
\hspace{2mm}+proto-rect &  en+fs-3.5 &  83.7 &85.2 &\textbf{83.5} &82.7 &84.1 &81.2 &\textbf{79.8} &\textbf{80.3} &81.2 &79.4 &\textbf{80.4} &\textbf{77.7} &\textbf{73.5} &\textbf{74.4}  & \textbf{80.5}\\
\hspace{2mm}+norm &  en+fs-3.5 &  83.1 &84.6 &82.8 &82.1 &83.5 &80.4 &79.0 &79.3 &80.4 &78.5 &79.6 &76.6 &71.8 &73.0  & 79.6\\
\hspace{4mm}+proto-rect &  en+fs-3.5 &  \textbf{83.8} &\textbf{85.2} & 83.4 &\textbf{82.8} &\textbf{84.2} &\textbf{81.3} &\textbf{79.8} & 80.2 &\textbf{81.3} &\textbf{79.4} & 80.3 &\textbf{77.7} & 73.2 & 74.2  & \textbf{80.5}\\
Fine-tuning (full) &  en+fs-3.5 &  83.2 &84.6 &82.9 &82.2 &83.5 &80.8 &79.2 &79.5 &80.5 &78.6 &80.2 &77.0 &72.6 &74.0  & 79.9\\
Fine-tuning (head) &  en+fs-3.5 &  83.2 &84.9 &83.2 &82.3 &83.5 &80.4 &79.0 &79.1 &80.3 &78.4 &79.6 &76.9 &72.9 &73.2  & 79.8\\
            \bottomrule
        \end{tabular}
    }
    \caption{Few-shot XNLI accuracy results across 14 languages with average improvements for each of the methods. All the confidence interval is less than .07 in the experiments. "fs-3.5" means 3-way-5-shot learning. }
    \label{table:xnli}
\end{table*}

\begin{table}[h!]
    \resizebox{1\linewidth}{!}{
        \begin{tabular}{lccccccccc}
        \toprule 
        Exp. Type & Resource & \textbf{de} & \textbf{es} & \textbf{fr} & \textbf{ja} & \textbf{ko} & \textbf{zh} & avg\\
        \midrule
        \midrule
        \multicolumn{9}{c}{$\theta_{\mathcal{T}_s}^{src}$  = Finetuned-XLM-R-large with PAWS-X dataset} \\
        \midrule
        \midrule
        Zero-Shot &  en &  89.8 &89.6 &90.5 &78.8 &78.6 &81.9  & 84.9\\
        NN &  en+fs-2.5 &  89.8 &89.8 &90.6 &79.8 &80.4 &82.5  & 85.5\\
        \hspace{2mm}+proto-rect &  en+fs-2.5 &  90.3 &90.2 &91.0 &80.5 &81.2 &83.3  & 86.1\\
        \hspace{2mm}+norm &  en+fs-2.5 &  90.0 &90.2 &90.8 &79.9 &80.7 &82.7  & 85.7\\
        \hspace{4mm}+proto-rect &  en+fs-2.5&  \textbf{90.4} &\textbf{90.6} &\textbf{91.2} &\textbf{80.5} &\textbf{81.3} &\textbf{83.5}  & \textbf{86.3}\\
        Fine-tuning (full) & en+fs-2.5 &  88.9 &89.1 &89.6 &79.2 &79.7 &82.0  & 84.7\\
        Fine-tuning (head) &  en+fs-2.5 &  90.0 &89.8 &90.7 &79.3 &79.5 &82.1  & 85.3\\
        \bottomrule
        \end{tabular}
    }
    \caption{Few-shot PAWS-X acc. results across 6 languages. Here in Resource column, "en" indicates model is trained with full English training data. \textit{fs-2.5} means 2-way-5-shot learning. }
    \label{table:pawsx_1}
\end{table}

\begin{table}[h!]
    \resizebox{1\linewidth}{!}{
        \begin{tabular}{lccccccccc}
        \toprule 
        Exp. Type & Resource & \textbf{en} & \textbf{de} & \textbf{es} & \textbf{fr} & \textbf{ja} & \textbf{ko} & \textbf{zh} & avg\\
        \midrule
        \midrule
        \multicolumn{9}{c}{$\theta_{\mathcal{T}_s}^{src}$  = Finetuned-XLM-R-large with XNLI dataset} \\
        \midrule
        \midrule
Zero-Shot &  en &  41.4 &43.5 &44.1 &43.8 &46.0 &46.7 &44.4  & 44.3\\
NN &  en+fs-2.5  &  71.5 &66.8 &65.2 &66.6 &60.1 &58.8 &61.8  & 64.4\\
\hspace{2mm}+proto-rect &  en+fs-2.5 &  70.5 &66.1 &65.1 &66.2 &60.0 &58.6 &61.6  & 64.0\\
\hspace{2mm}+norm &  en+fs-2.5 &  \textbf{72.2} &\textbf{67.8} &\textbf{66.1} &\textbf{67.2} &\textbf{60.8} &\textbf{59.7} &\textbf{62.5}  & \textbf{65.2}\\
\hspace{4mm}+proto-rect & en+fs-2.5 &  72.0 &67.5 &65.9 &66.7 &61.0 &59.5 &62.8  & 65.0\\
Fine-tuning (full) &  en+fs-2.5 &  64.4 &59.4 &58.3 &59.6 &54.0 &53.7 &54.8  & 57.7\\
Fine-tuning (head) &  en+fs-2.5 &  48.2 &47.9 &48.3 &48.2 &47.7 &48.2 &46.8  & 47.9\\
        \bottomrule
        \end{tabular}
    }
    \caption{PAWS-X accuracy results for cross-task experiments across 6 languages. For this experiment, we fine-tuned XLM-R LM using the XNLI task and then applied few-shot inference on the PAWS-X task.}
    \label{table:pawsx_2}
\end{table}


\subsection{Fine-tuning}
We use XLMR-large \cite{XLMR} as our  pre-trained language model $\theta_{lm}$ and perform standard fine-tuning using labeled English data to adapt it to task model $\theta_{\mathcal{T}_s}^{src}$. We tune the hyper-parameters using English development data and report results using the best performing model (optimal hyper-parameters have been enlisted in the appendix). We train our model using 5 different seeds and report average results across them. We use the same optimal hyper-parameters to fine-tune on the target languages. As baseline we add two additional fine-tuning named \textit{head} and \textit{full}. Fine-tuning \textit{full} means all the parameters of the model are updated. This is very unlikely in Few-shot scenarios.  Fine-tuning head means only the parameters of the last linear layer are updated.

\subsection{Evaluation Setup}\label{sec:eval_setup}
\citet{nooralahzadeh2020zeroshot} and \citet{lauscher2020zero} used 10 and 5 different seeds to measure the few-shot performance. As few-shot learning involves randomly selecting small support sets, results may vary greatly from one experiment to the next, and hence may not be reliable \cite{le2020continual}. In computer vision, \textit{episodic testing} \cite{Ravi2017OptimizationAA,DBLP:journals/corr/abs-1905-11116,ziko2020laplacian} is often used for evaluating few-shot experiments. Each episode is composed of small randomly selected support and query sets. Model's performance on each episode is noted, and the average performance score, alongside the  confidence interval (95\%) across all episodes are reported. To the best of our knowledge, episodic testing has not been leveraged for cross-lingual few-shot learning in NLP.

\noindent We evaluate our approach using 300 episodes per seed model $\theta_{\mathcal{T}_s}^{src}$ totalling 1500 episodic  testing and report their average scores. For each episode, we perform C-way-N-shot inference.  For  2-way-5-shot setting, for instance, we randomly select 15   query samples per class, and $2 \times 5$ number of support samples. For XNLI and PAWS-X, we use $3$ and $2$ as the value of C, respectively. Our episodic testing approach has been detailed further in the Episodic Algorithm  of the Appendix.

\subsection{Results and Analysis}  
\label{sec:results}

After training the model with the source language samples (i.e. labeled English data), we perform additional fine-tuning using $C$-way-5-shot target language samples. Finally, we perform our proposed NNFS inference.

\noindent The fine-tuning baseline using limited target language samples result in small but non-significant improvements over the zero-shot baseline. The NNFS inference approach, however, resulted in  performance gains using only 15 (3-way-5-shot) and 10 (2-way-5-shot) support examples for both XNLI and PAWS-X tasks. When compared to the few-shot baseline, we got an average improvement of 0.6 on XNLI (table \ref{table:xnli}) and 1.0 on PAWS-X (table \ref{table:pawsx_1}). At first we experimented with 3-shot support samples but did not observe any few-shot capability in the model. We also experimented with 10-shot setup and found similar improvements of NNFS on top of the Fine-tuning baseline (results have been added to the  Appendix).  Interestingly, for both cases, we observed higher performance gains on low resource languages.

\noindent To further evaluate the effectiveness of our model, we tested it in a cross-task setting. We first trained the model on XNLI (EN data) and then used NNFS inference on PAWS-X. Table  \ref{table:pawsx_2} demonstrates an impressive average performance gain of +7.3 across all PAWS-X languages, over the fine-tuning baseline. 



\noindent In addition to that, NNFS inference approach is fast. When compared to the zero-shot inference ($1 \textsc{x}$), our approach takes only $\approx 1.36-1.7 \textsc{x}$ time of computation cost compared to the finetuning time which takes $\approx 38-40 \textsc{x}$. Table 6 in appendix shows the inference time details on both  tasks.




\vspace{-.5em}
\section{Conclusion}
\vspace{-.5em}
The paper proposes a nearest neighbour based few-shot adaptation algorithm accompanied by a necessary evaluation protocol. Our approach does not require updating the LM weights and thus avoids over-fitting to limited samples. We experiment using two classification tasks and results demonstrate consistent improvements over finetuning not only across languages, but also across tasks.

\bibliography{ref.bib}
\bibliographystyle{acl_natbib}

\newpage
\clearpage
\appendix
\section{Appendices}
\label{sec:appendix}

\subsection{Decision choice for Episodic Testing}
In the traditional testing framework, we sample a batch from the dataset and calculate the batch's prediction. Finally, accumulate all the predictions to calculate the score of the evaluation metric. However, Few-shot experiments are quite unpredictable because of the following two reasons,

\begin{itemize}
    \item Support set: Per class sampling strategy of the support set is random. In a few shot experiments, we perform inference on the test dataset utilizing support-samples. For a different support set, the prediction may vary drastically. However, taking few samples (ie., 10 out of 2500 or 15 out of 2000) and doing experiments 5-10 times doesn't reflect the true potential of a few-shot algorithm. 
    \begin{figure}[!ht]
        \centering
        \scalebox{1}{
          \includegraphics[scale=.44]{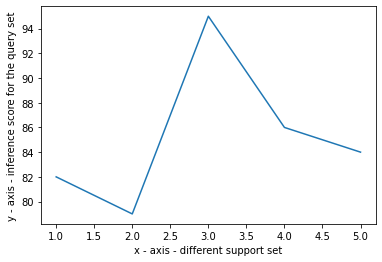}}
        \caption{For a same query set result varies because of different support set.}  
        \label{fig:episodic-test}
    \end{figure}

    \item Transductive inference: On the contrary, for a few shot experiments, algorithms often perform transductive inference. In transductive inference, predictions may vary based on the combination of the query samples. Hence it is challenging to benchmark the few shot algorithms with the traditional testing framework. 
\end{itemize}	

\noindent In Episodic testing, we randomly sample a query set and support set from the dataset and perform few-shot experiments.  We perform the experiments until we get a low confidence-interval (95\%). In this way, we may iterate over the test dataset 5-10 times more. However, it is not affected by the above problems mentioned and can benchmark any few-shot algorithm properly.

\begin{figure*}[!ht]
    \centering
    \scalebox{1}{
      \includegraphics[scale=.44]{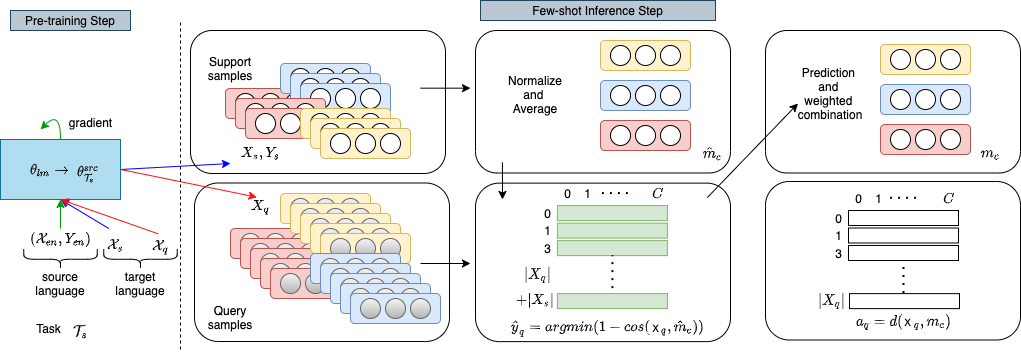}}
    \caption{Training flow diagram of nearest neighbour few-shot learning for cross-lingual NLP. In the  \textbf{Pre-training Step} we train a language model $\theta_{lm}$ on the source language (en) data $(\mathcal{X}_{src}, \mathcal{Y}_{src})$ to get $\theta_{\mathcal{T}_s}^{src}$. In \textbf{Few-Shot Inference Step}, we apply forward propagation on the $\theta_{\mathcal{T}_s}^{src}$ model using support input samples $\mathcal{X}_s$ and $\mathcal{X}_q$ and get the latent representations  $X_s$ and $Y_s$. Using $X_s$, we apply normalization and calculate $m_c$. We then use both $X_s$ and $X_q$, and compute the unary term $a_q$, which in turn gives the label distribution of the query samples (see in Alg. Few Shot Inference. line \#14-15 ). }  
    \label{fig:proposed-model}
\end{figure*}

\begin{algorithm*}[t!]
    \footnotesize
    \caption{Episodic Testing}
    \label{alg:FewShotAdaptationTesting}
     \textbf{Input:} Model $\theta_{\mathcal{T}_s}^{src}$ trained using the source language, transductive parameter $\lambda$, mean representation of train/dev samples $m_s$, a threshold value $eps$, a multiplier $\tau$, input data ($C$-way-$N$-shot) \\
     \textbf{Output:} Average $\bar{s}$ score and the  confidence interval $\partial$  
    \begin{algorithmic}[1]
        \setstretch{1.2}
        \State $s_e = \phi$ \algorithmiccomment{score list for all the episodes.}
        \For{$e \in [1:totEpisode]$}
            \State \texttt{/* Random sampling */}
            \State \texttt{/* Randomly select $C$ number of classes */}
            \State MetaClass = SelectRandomClasses($C$)
            \State \texttt{/*bs = Batch Size*/} 
            \State $\mathcal{X}_s, \mathcal{Y}_s$ = SupportIterator.next(bs=($C$-way-$N$ shot)) \algorithmiccomment{Iterator $\in$  MetaClass.} 
            \State $\mathcal{X}_q$ = QueryIterator.next(bs=($C$-way-$N \times \tau$ shot)) \algorithmiccomment{Iterator $\in$  MetaClass.}
            \State \texttt{/* Perform Inference using Respective few-shot algorithm. */}
            \State $\mathcal{Y}_q$ = FewShotInference($\theta_{\mathcal{T}_s}^{src}, (\mathcal{X}_s, \mathcal{Y}_s), (\mathcal{X}_q), \lambda, m_s, eps$) \algorithmiccomment{Using Alg. \emph{Few Shot Inference}.}
            \State $s$ = CalcScore($\mathcal{Y}_q$)
            \State $s_e \ \ = \ \ s_e \ \ \cup \ \ \{s\}$
        \EndFor
        \State $\bar{s}$, $\partial$ = Average($s_e$), ConfidenceInterval($s_e$)
        \State Return $\bar{s}$, $\partial$
        
    \end{algorithmic}
    \end{algorithm*}

\subsection{Extended Dataset}\label{app_sec:ext_dataset}

\paragraph{XNLI} We use XNLI dataset \cite{XNLI} which extends the MultiNLI dataset \cite{multiNLI} to 15 languages. MultiNLI dataset contains sentences from 10 different genres. The objective is to identify if a \emph{premise} entails with the \emph{hypothesis}. It is a crowd sourced 3-class classification dataset covering 14 languages that have been translated from English. These locales include French (fr), Spanish (es), German (de), Greek
(el), Bulgarian (bg), Russian (ru), Turkish (tr),
Arabic (a), Vietnamese (vi), Thai (th), Chinese
(zh), Hindi (hi), Swahili (sw), and Urdu (ur). It comes with human translated  dev and test splits. The dataset is balanced and contains 392702, 2490 and 5010 numbers of train, dev and test instance for each of the language, respectively.

\paragraph{PAWSX} Given a pair of sentences, the objective of PAWS (Paraphrase Adversaries from Word Scrambling) \citep{zhang-etal-2019-paws} is to classify  if the pair is a paraphrase or not. PAWS-X dataset contains six topologically different languages that have been machine translated from English. These include French (fr), Spanish (es), German (de), Korean (ko), Japanese (ja), and Chinese (zh). Similar to XNLI, it also comes with  human translated  dev and test split.

\paragraph{Challenges} Both datasets posses different challenges. NLI task requires rich and a high level of factual understanding of the text. The PAWS task, on the other hand, contains pairs of sentences that usually have a high lexical overlap and may/may not be paraphrases. We use \textit{accuracy} as the evaluation metric for both datasets.

\paragraph{10 Shot results} 
For reference we have added 10 shot experiment for XNLI and PAWSX dataset with same setup as Table 1 and Table 2 of main paper.

\begin{table*}[h!]
    \resizebox{1\linewidth}{!}{%
        \begin{tabular}{lcccccccccccccccc}
            \toprule 
            Exp. Type & Resource & \textbf{fr} & \textbf{es} & \textbf{de} & \textbf{el} & \textbf{bg} & \textbf{ru} & \textbf{tr} & \textbf{ar} & \textbf{vi} & \textbf{th} & \textbf{zh} & \textbf{hi} & \textbf{sw} & \textbf{ur} & avg\\
            \midrule
            \midrule
            \multicolumn{17}{c}{$\theta_{\mathcal{T}_s}^{src}$  = Finetuned-XLM-R-large with XNLI dataset} \\
            \midrule
            \midrule
            
Zero-Shot &  en &  83.1 &84.8 &83.0 &82.1 &83.3 &80.2 &78.9 &78.7 &80.1 &78.1 &79.5 &76.7 &72.5 &73.0  & 79.6\\
NN &  en+fs-3.10 &  83.4 &85.0 &83.1 &82.5 &83.8 &80.9 &79.5 &79.8 &80.8 &79.2 &80.3 &77.4 &72.9 &74.0  & 80.2\\
NN+proto-rect &  en+fs-3.10 &  \textbf{83.8} &\textbf{85.3} &\textbf{83.6} &\textbf{82.8} &84.1 &81.2 &79.9 &\textbf{80.4} &81.3 &79.6 &\textbf{80.7} &\textbf{78.1} &\textbf{73.6} &\textbf{74.7}  & 80.6\\
NN+norm &  en+fs-3.10 &  83.5 &85.0 &83.2 &82.6 &83.8 &81.1 &79.6 &79.8 &81.0 &79.3 &80.3 &77.5 &73.0 &74.0  & 80.3\\
NN+norm+proto-rect &  en+fs-3.10 &  \textbf{83.8} &85.2 &\textbf{83.6} &\textbf{82.8} &\textbf{84.2} &\textbf{81.4} &\textbf{80.0} &\textbf{80.4} &\textbf{81.4} &\textbf{79.7} &\textbf{80.7} &\textbf{78.1} &73.5 &74.6  & \textbf{80.7}\\
Fine-tuning (full) &  en+fs-3.10 &  83.2 &84.5 &82.9 &82.5 &83.7 &81.2 &79.5 &79.8 &80.8 &78.9 &80.5 &77.3 &72.6 &74.2  & 80.1\\
Fine-tuning (head) &  en+fs-3.10 &  83.3 &85.0 &83.2 &82.4 &83.5 &80.6 &79.2 &79.4 &80.5 &78.6 &79.9 &77.2 &72.8 &73.6  & 79.9\\
            \bottomrule
        \end{tabular}
    }
    \caption{10-shot XNLI accuracy results across 14 languages with average improvements for each of the methods. All the confidence interval is less than .07 in the experiments.}
    \label{table:xnli}
\end{table*}

\begin{table*}[h!]
    \centering
    \resizebox{.7\linewidth}{!}{
        \begin{tabular}{lccccccccc}
        \toprule 
        Exp. Type & Resource & \textbf{de} & \textbf{es} & \textbf{fr} & \textbf{ja} & \textbf{ko} & \textbf{zh} & avg\\
        \midrule
        \midrule
        \multicolumn{9}{c}{$\theta_{\mathcal{T}_s}^{src}$  = Finetuned-XLM-R-large with PAWS-X dataset} \\
        \midrule
        \midrule
        Zero-Shot &  en &  89.8 &89.6 &90.6 &78.8 &78.4 &81.8  & 84.8\\
NN &  en+fs-2.10 &  90.0 &90.1 &90.8 &80.2 &80.7 &83.2  & 85.8\\
NN+proto-rect &  en+fs-2.10 &  90.3 &90.3 &91.2 &80.5 &81.2 &83.5  & 86.2\\
NN+norm &  en+fs-2.10 &  90.1 &90.4 &91.1 &80.3 &81.0 &83.3  & 86.0\\
NN+norm+proto-rect &  en+fs-2.10 &  \textbf{90.4} &\textbf{90.7} &\textbf{91.4} &\textbf{80.7} &\textbf{81.5} &\textbf{83.7 } & \textbf{86.4}\\
Fine-tuning (full) &  en+fs-2.10 &  89.4 &89.6 &90.1 &79.9 &80.6 &82.6  & 85.4\\
Fine-tuning (head) &  een+fs-2.10 &  90.1 &90.1 &91.0 &79.9 &79.8 &82.5  & 85.6\\
        \bottomrule
        \end{tabular}
    }
    \caption{10-shot PAWS-X acc. results across 6 languages. Here in Resource column, "en" indicates model is trained with full english training data.}
    \label{table:pawsx_1}
\end{table*}

\subsection{Hyperparameters and Resource Description}\label{app_sec:hyperparams}

We used 8 V100 GPUs (amazon \emph{p3.16xlarge}) to run all experiments. The hyper-parameters of the best performing model are enlisted in Table \ref{table:hyp}. In the pretrained language model finetuning, We use (1e-5, 3e-5, 5e-5, 7.5e-6 ,5e-6) boundary values to search for proper learning rate.

\begin{table*}[!ht]
\begin{center}
\begin{tabular}{l|c}
\hline 
\bf Hyperparameter & \bf Value \\  
\hline
LM & XLMR-large\\
\# of params & 550M\\
learning rate & 7.5e-6\\
Max Sequence Length & 128 \\
Per GPU batch size & 8\\
Gradient accumulation step & 2\\
Multi-GPU training & 8\\
Effective batch size & 128\\
Number of epoch & 10\\
Warmup step in pre-training & 6\% of total number of steps\\
Total number of episodic test & 1000\\
finetuning batch-size & 16\\
finetuning learning rate & 7.5e-6\\
finetuning schedueler & constant scheduler\\
\hline
\end{tabular}
\end{center}
\caption{Optimal hyper-parameter settings.}
\label{table:hyp}
\end{table*}

\begin{table*}[h!]
    \centering
    \resizebox{.6\linewidth}{!}{
        
        \begin{tabular}{lccccc}
        \toprule 
        Exp. Type & \multicolumn{2}{c}{PAWSX} & \multicolumn{2}{c}{XNLI}\\
         & fs-2.5 & fs-2.10 & fs-3.5 & fs-3.10 \\
        \midrule
        Zero-Shot & 1x & 1x & 1.35x & 1x\\
        NN & 1.36x & 1.71x & 1.35x & 1.66x  \\
        \hspace{2mm}+proto-rect & 1.37x & 1.71x & 1.35x & 1.67x\\
        \hspace{2mm}+norm & 1.36x & 1.71x & 1.35x & 1.66x\\
        \hspace{4mm}+proto-rect & \textbf{1.37x} & \textbf{1.71x} & \textbf{1.35x} & \textbf{1.67x}\\
        Fine-tuning (full) & 22.44x & 41.86x & 21.01x & 38.69\\
        Fine-tuning (head) & 20.48x & 38.02x & 19.24x & 35.17\\
        \bottomrule
        \end{tabular}
    }
    \caption{Inference time for each of the task.}
    \label{table:time}
\end{table*}

\section{Reproducibility Settings and Notes}

\begin{itemize}
    \itemsep0em 
    \item $python_{3.6.13}$. $Pytorch_{1.7.1}$, $CUDA_{10.2}$, $cuDNN_{7605}$
    \item $transformers_{4.6.0}$
    \item Average runtime: See table 7.
\end{itemize}

\end{document}